# Design and Implementation of Thermal Control Strategy for Micro-size Lunar Exploration Rover HAKUTO

**Toshiki Tanaka[a], Takuto Oikawa[b], Keeni Shruti[b], Teruhito Iida[a], Prof. Kazuya Yoshida[b], Dr. John Walker[a]**

[a] *ispace inc. Sumitomo Shibakoen Building 10F, 2-7-17, Shiba, Minato-ku, Tokyo, Japan, https://ispace-inc.com/*
[b] *The Space Robotics Lab, Tohoku University, Japan http://www.astro.mech.tohoku.ac.jp/e/*

**Abstract**

The research focuses on the thermal interface design of the HAKUTO Rover Flight Model (FM) and its evaluation using the final thermal vacuum experiment. Initially, an analytical thermal model was developed, dividing the rover into six distinct radiating surfaces. The surface temperature of the rover was then determined based on its orientation relative to the solar radiative flux, as well as other thermal factors such as lunar radiation, solar albedo, and conduction from the lunar regolith. By considering the impact of these heat sources, suitable mounting locations for the avionics and the angle of the solar panels were determined to optimize the rover's thermal performance and power generation while ensuring the rover's mass remained under 4 kg. After the preliminary thermal analysis, numerical computational simulations were conducted to further elaborate on the thermal response of the final HAKUTO Rover FM design, which was then verified through a thermal balance test in a vacuum environment.

**Nomenclature**

$Ai$: Surface area
$b$: Surface albedo
$F_{l,\,i}$: View factor from the lunar surface to the rover
$F_{s,\,i}$: View factor from sun to the rover
$Ps$: Solar irradiance
$T_{r,\,i}$: Rover's temperature
$T_{sp}$: Outer space temperature
$\alpha_i$: Rover's surface Solar absorptivity
$\varepsilon_i$: Rover's surface emissivity
$\varepsilon_l$: Lunar surface emissivity
$\lambda$: Thermal conductivity
$\sigma$: Stefan-Boltzmann constant

## 1. Introduction

Recently, there has been an increase in space missions led by private ventures. The Google Lunar X Prize (GLXP) [1] was a competition intended to incentivize the development of low-cost access to the lunar surface with minimal aid from government funds. The competition ended at the end of March 2018 without a winner; however, several companies gained valuable business experiences and successfully developed hardware capabilities in the process. Our team, Team HAKUTO, was selected as one of the five finalists and has successfully completed the development of a lunar micro-rover Flight Model (FM). Team HAKUTO has also developed a unique thermal control system to enable the use of commercial off-the-shelf hardware in the extreme thermal environment of the Moon while ensuring the rover remains under 4 kg in mass.

This paper focuses on the thermal design of the HAKUTO Rover, as depicted in Fig. 1, and its evaluation using the final thermal balance experiment in a vacuum chamber. In previous research, thermal vacuum experiments were conducted on a Pre-FM (PFM) prototype to verify the thermal behavior of the proposed design in preparation for FM development [2, 3]. Based on the results of these earlier experiments, we identified necessary improvements to ensure critical components remain within their operational temperature limits for the expected thermal environment.

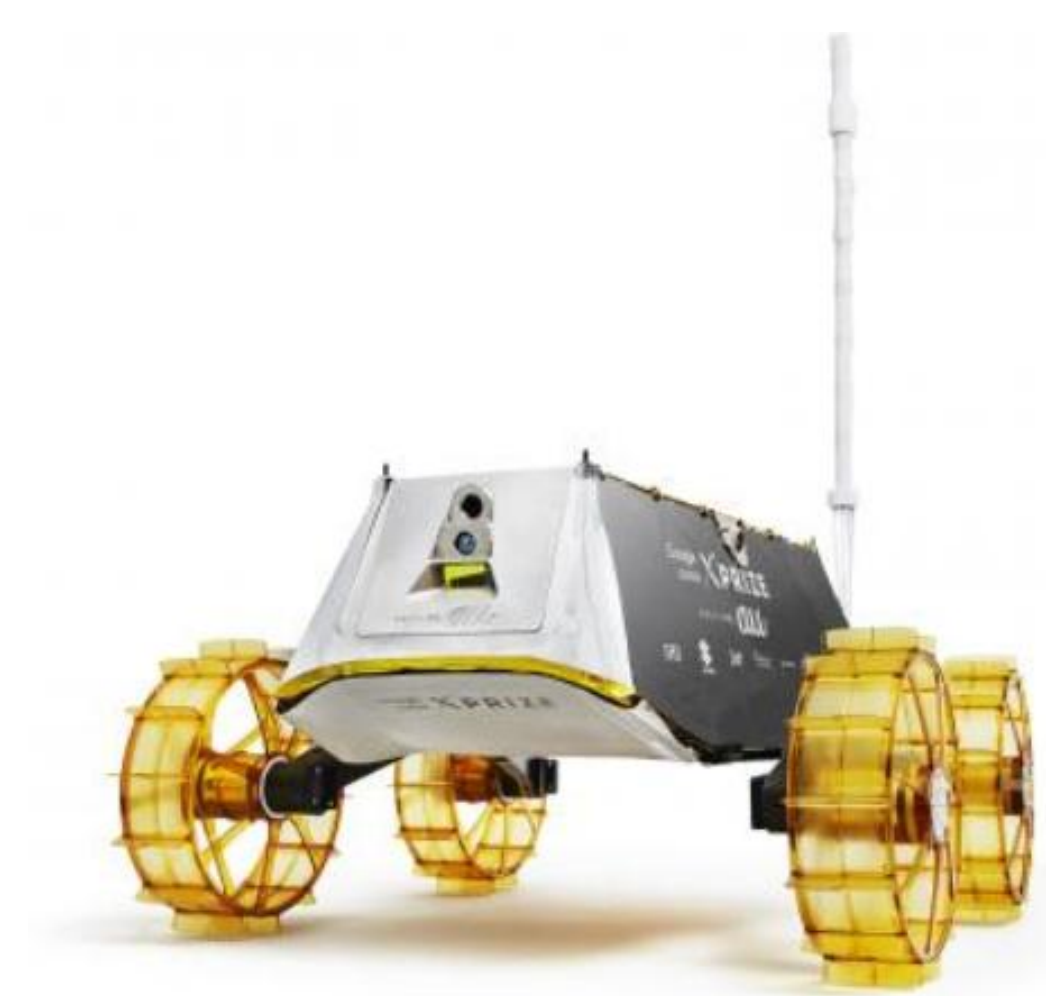

Fig. 1. HAKUTO Flight Model Rover

### *1.1 Key Challenges*

The most significant constraints of the HAKUTO rover's thermal design were mass and size. The cost for launching a rocket is calculated in proportion to the mass of the rover. Therefore, mass reduction was crucial for maintaining a balance in the financial aspects of our business plan. Based on our business plan, the engineering and business teams made the decision to limit the rover's mass to less than 4 kg including an interface between the lander and the rover.

*1.2 Thermal Environment*

The HAKUTO mission timeline was divided into seven phases: ground, launch, cruise, thrusting, waiting on the Moon, deployment, and surface mission phases, as shown in Table 1. During the ground, launch, and waiting on the Moon phases, the rover will be kept in a thermally controlled environment with moderate temperatures. Therefore, our analysis focused on the cruise, thrusting, deployment, and surface mission phases.

Table 1. Mission Phase Definition

| Phase | Remarks |
|---|---|
| Ground Phase | On the Earth, before launch |
| Launch Phase | During launch, in a rocket fairing |
| Cruise Phase | After separation from the rocket, cruising from the Earth to the Moon |
| Thrusting Phase | During thrusting upon landing sequence |
| Waiting on the Moon Phase | After landing until the deployment on the Moon. In this phase, with the help of heater provided by a lander, the temperature of the rover will be kept at 20°C. |
| Deployment Phase | After deployment in the shadow area until getting out of the shadow area. |
| Surface Mission Phase | After getting out of the shadow of the lander until the end for the mission. |

Cruise Phase

In this phase, the HAKUTO rover will be attached to the lander spacecraft developed by Team Indus. The rover will be directly mounted onto the lander deck using a Frangibolt, which will be released to deploy the rover onto the Moon. While attached to the lander deck, the rover is partially encapsulated by a box-shaped cover called the Lander interface, which is designed by Team HAKUTO, as depicted in Fig. 2.

Due to the severe mass constraint, we decided against fully enclosing the rover in a box. Instead, we extended four legs from the lander and placed multi-layer insulation (MLI) around them to reduce heat exchange between the rover and the lander, as well as between the rover and outer space. Based on the interface design agreed upon between Team Indus and Team HAKUTO, the temperature of the lander deck will be maintained above -50°C. Additionally, internal heaters will be used to warm the rover. The heaters and corresponding temperature sensors will be designed by Team HAKUTO but monitored and controlled by Team Indus. Due to the significant limitations in rover size and power, Team HAKUTO thoroughly investigated the layout of the heaters, aiming to minimize heater power while adhering to the design constraints.

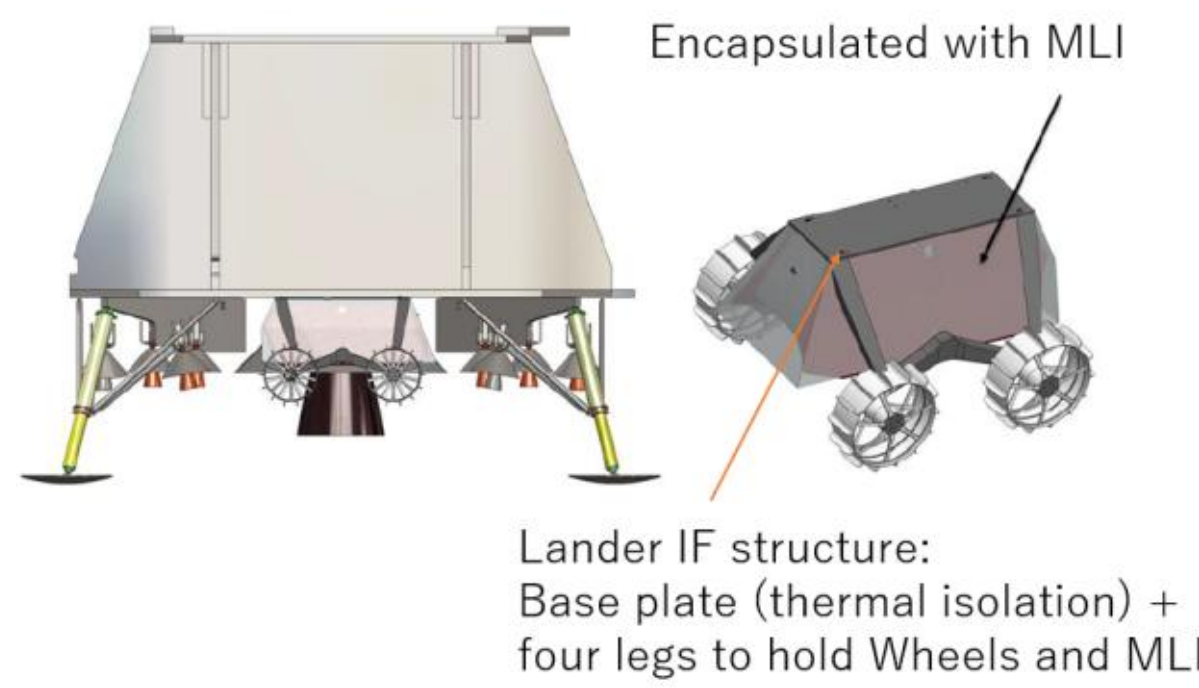

Fig. 2. Conceptual image of the Lander Interface

Thrusting Phase

During this phase, the main thruster of the lander is activated, creating an exhaust plume that will last for up to 15 minutes. This exhaust plume generates a significant heat load on the rover. To assess the impact on the rover's thermal performance, we conducted thermal simulations meticulously. The results of these simulations have provided confirmation that the rover's temperature will remain within its designated thermal range throughout the duration of thrust operations.

Deployment Phase

Until the rover is deployed, the temperature of the lander deck will be actively controlled, ensuring that the rover remains within its designated operating temperature range. However, once the rover is detached from the lander and deployed onto the Moon, there is a possibility that it may end up in a shadow zone, with the lander blocking the sunlight. In such a scenario, the rover will be unable to receive any energy from the sun for power generation. During this phase, it becomes crucial to assess the duration for which the rover might remain in shadow and determine how soon it will need to leave these shadowed regions. This assessment is vital to ensure the rover's power is maintained and its operations will not be compromised.

Surface Operation Phase

Surface temperature profiles at various latitudes are shown in Fig. 3, using data from the Diviner Lunar

Radiometer Experiment (Diviner) [4]. A lunar period lasts approximately 28 Earth days, encompassing both lunar day and night, spanning about two Earth weeks. As the rover relies solely on solar panels for power generation [3], the mission is designed to operate within a single lunar day.

The timing of the mission start is crucial due to the extreme temperature conditions during lunar sunrise, where lunar temperatures can drop to -200°C. To mitigate the risk of failures caused by these low temperatures, we have determined that the mission should commence no earlier than 50 Earth hours after lunar sunrise, allowing the surface temperature to reach at least -20°C in the target region at 28° latitude. During lunar noon, the rover will experience the maximum heat from the sun and the lunar surface [3]. According to the profile of the lunar surface temperature, we have made the decision to establish a mission period during which the rover's temperature should be maintained within its designated operating temperature range. This mission period is defined as starting 50 hours after the local dawn and lasting until 50 hours before the local sunset.

Based on the Diviner data, Hurley et al. [5] have successfully developed a mathematical model to predict lunar surface temperatures. In the case of the lunar day condition, the temperature model (Eq1) is defined as a function of the solar zenith angle between the Moon-Sunline and the vector from the center of the Moon to the specific point on the lunar surface ($\psi$). The tuning factor (n) is incorporated to match the temperature profile in sunlit conditions. Furthermore, the one-day lunar revolution at specific latitude ($\phi_{lat}$) can be rearranged into Earth-time (t) reference, where Oikawa et al. has defined it as following [3]:

$$T_l(\psi) = 392cos^n(\psi) > 130K \quad (\psi < 90°) \tag{1}$$

$$\psi(t) = \frac{2\pi}{T_p}\left|t - \frac{T_p}{4}\right|cos(\phi_{lat} + \phi_{inc}) \tag{2}$$

This equation is true if the temperature does not fall below a threshold of 130°K. Moreover, $\psi$ is rearranged into a time-variant function ($t$) to calculate the time of day relative to the sun's position. Parameters such as the surface temperature at a target lunar latitude ($\phi_{lat}$), overall lunar inclination angle ($\phi_{inc}$), and lunar revolution period ($Tp$) are introduced to formulate this relationship. For the n power factor, we have changed the correction factor to 0.29 as proposed by Walker [6] to reduce the daytime temperature model error to within 10°C when compared to the Diviner data instead of a correction factor of 0.25 by Hurley et al [5].

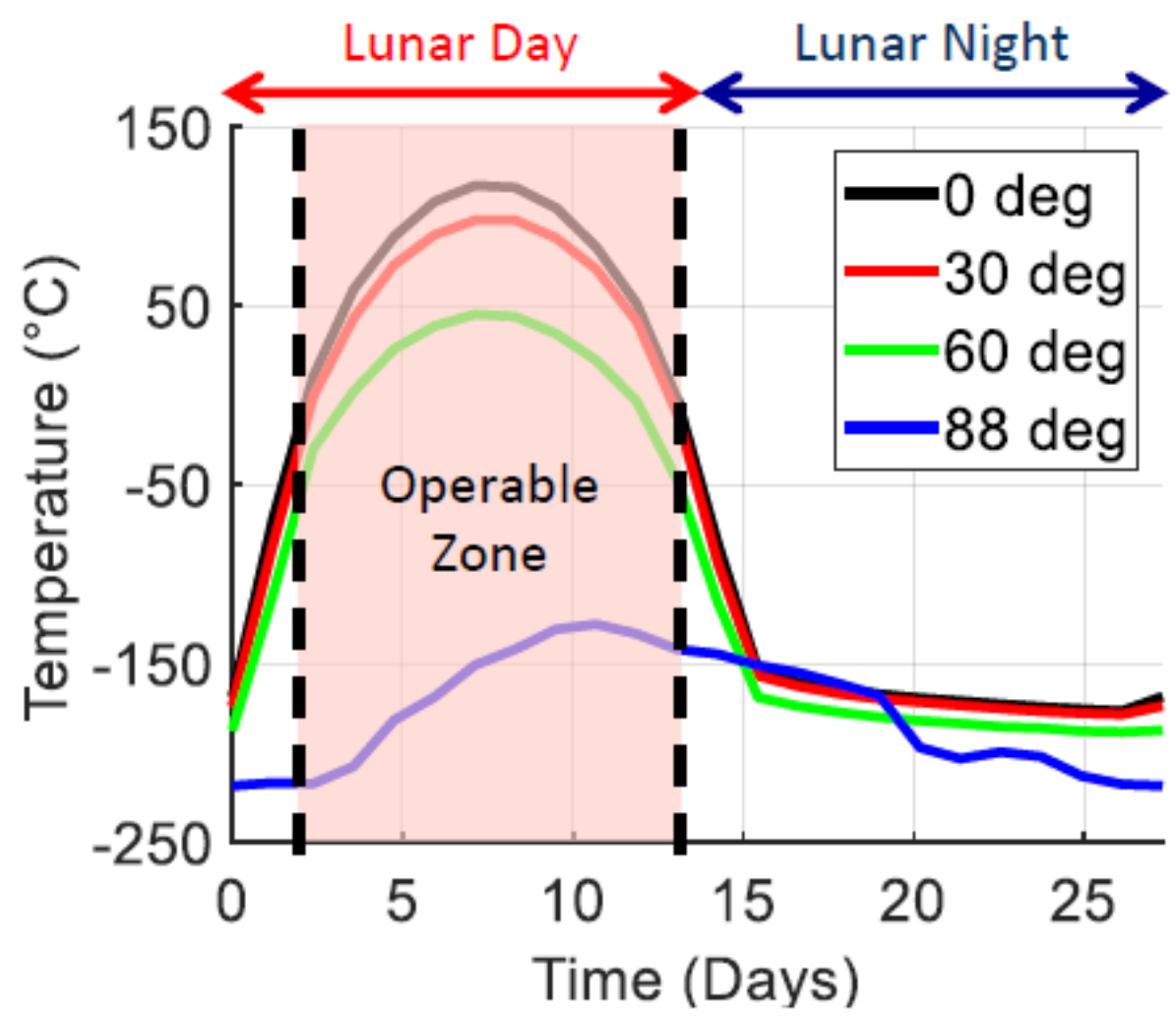

Fig. 3. Lunar Surface Temperature Model

*1.3 Design Requirements*

The goal of thermal systems design is to ensure that all components remain within their designated operating temperature ranges throughout all operational phases. The operational and storage temperature requirements for each component are provided in Table 2.

Table 2. Operation/Storage Temperature Range of Installed Components

| Components | Operation Temperature Range [°C] | Storage Temperature Range [°C] |
|---|---|---|
| Battery | −5 – +45 | −20 – +60 |
| Radiation Sensor | −30 – +60 | −30 – +60 |
| CPU | −45 – +80 | −45 – +85 |
| Power Distribution Unit (PDU) | −40 – +85 | −40 – +85 |
| Radio | −10 – +80 | −10 – +80 |
| Motor Controller | −40 – +85 | −40 – +85 |
| Time of Flight (ToF) | −40 – +80 | −40 – +85 |
| Front Camera | −25 – +80 | −40 – +85 |
| Rear Camera | −30 – +70 | −40 – +85 |
| Right Camera | −25 – +80 | −40 – +85 |
| Left Camera | −30 – +70 | −40 – +85 |
| Motor | −40 – +100 | −165 – +100 |
| Solar Panel | −40 – +120 | −80 – +120 |

## 2. Design Approach and Resulting Design

A thermal design approach introduced by E.W. Hammond [7] was adopted as a guideline to ensure clear and concise trade-offs in the design process. In this section, we focus on the thermal analysis of the surface mission phase, as it serves as the primary driver for the thermal design of the rover.

*2.1 Fundamental Analysis Model*

Figure 4 illustrates the representative heat inputs and outputs originating from both the lunar environment and the inner body of the rover. In the diagram, $Q_s$ represents the solar radiation from the sun, $Q_a$ signifies the radiation resulting from the surface albedo effect, $Q_{lr}$ denotes the lunar surface radiation, $Q_c$ represents the conduction between the lunar surface and the rover, $Q_e$ signifies the heat generated from on-board electronics, and $Q_{sp}$ indicates the radiation emitted back to outer space. Assuming that each surface has a specific nodal point denoted as $i$, the following relation is derived to determine the general thermal equations.

$$delta\ Q_i = Q_{s,i} + Q_{\alpha,i} + Q_{lr,i} + Q_c + Q_{e,i} - Q_{sp,i} \quad (3)$$

where *delta Qi* = 0 in steady state.

Each heat transfer component is defined by the following equations.

$$Q_{s,i} = \alpha_i F_{s,i} A_i P_s \quad (4)$$

$$Q_{a,i} = bQ_{s,i} \quad (5)$$

$$Q_{lr,i} = \epsilon_i \epsilon_l F_{l,i} A_i \sigma (T_l^4 - T_{r,i}^4) \quad (6)$$

$$Q_c = \lambda \frac{A_w}{L_w} (T_r - T_w) \quad (7)$$

$$Q_e = constant \quad (8)$$

$$Q_{sp,i} = \epsilon A F_{sp,i} \sigma (T_{r,i}^4 - T_{sp}^4) \quad (9)$$

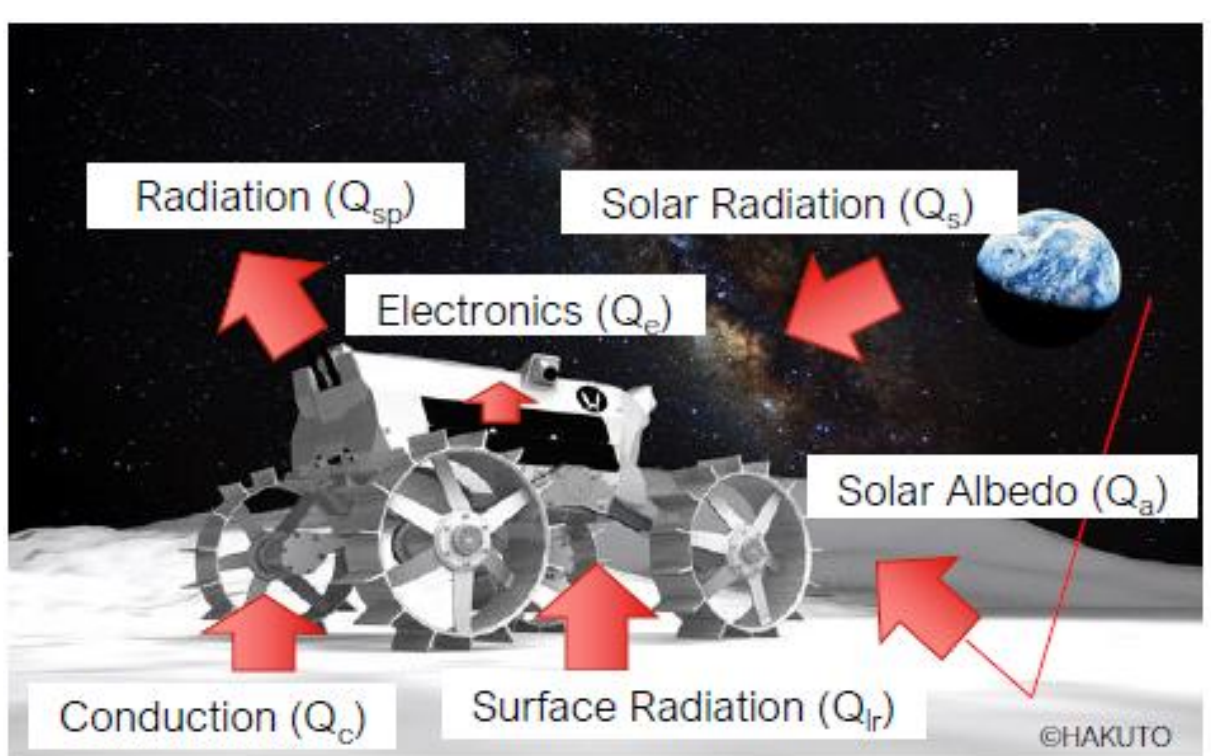

Fig. 4. External and internal heat flow on the lunar surface.

*2.2 Design Concept*

We implemented isolation between the body panels of the rover to simplify the thermal design process. By reducing the number of thermal interactions to consider, we can more easily assess and validate trade-offs.

Initially, we divided the rover into six distinct thermal zones: top-rear-front, right, left, bottom, as well as the wheels on the right and left side, as depicted in Fig. 5. To achieve isolation between these zones, we incorporated two thermal isolation measures. Firstly, we utilized a high thermal resistance material to minimize thermal conduction between the zones. Secondly, we employed multi-layer insulation (MLI) to reduce radiation.

This approach allows us to treat each node independently during the preliminary design phase, as they do not have thermal interaction with one another. Assuming a steady-state condition, the temperature of each node was calculated separately, as depicted in Fig. 6. This isolation concept provides a solid starting point and enables faster design iterations in the preliminary design phase, allowing for efficient thermal analysis and optimization.

In our design, we strategically placed all of the electronics on the top, rear, and front surfaces of the rover, as depicted in Fig. 7, considering the fact that these surfaces have smaller view factors to the lunar surface. The top surface, in particular, does not have a direct view factor to the lunar surface. This makes it relatively easier to design its heat balance by considering only the solar radiation ($Q_s$), heat generated from on-board electronics ($Q_e$), and radiation emitted back to outer space ($Q_{sp}$).

To further optimize the thermal design, front, rear, and side surfaces were oriented upward to avoid the lunar radiation factors. On the other hand, bottom surface is inevitably oriented towards the lunar surface. As a result, it is thermally coupled with the lunar surface's temperature, which can vary significantly throughout the mission period. Due to this coupling, the bottom surface is considered the least favorable place to position thermal sensitive items on the rover.

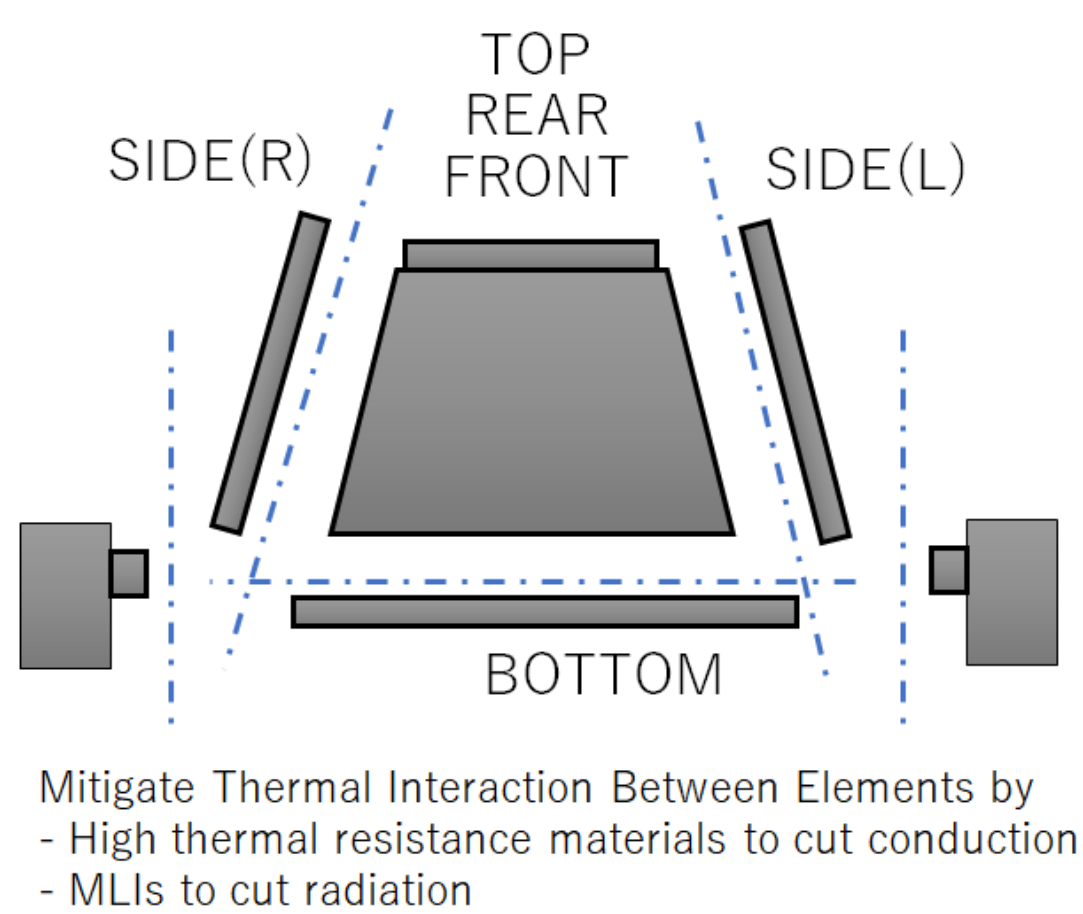

Fig. 5. Isolation concept

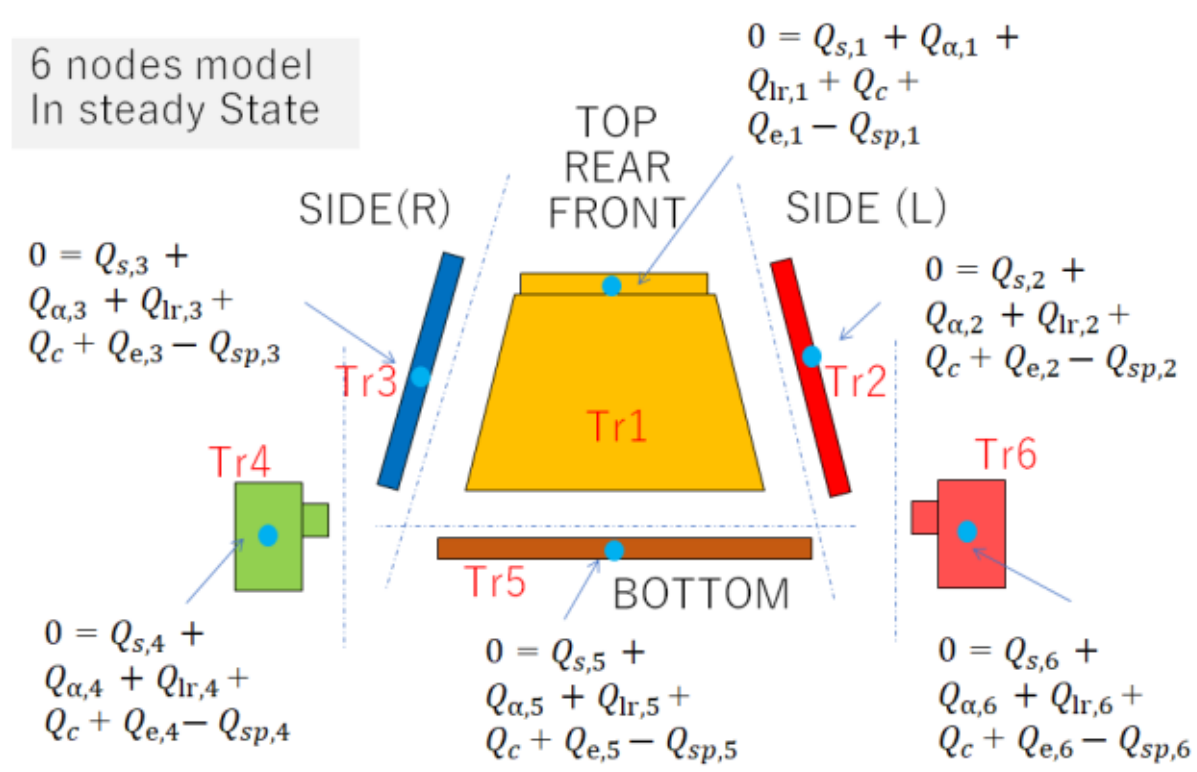

Fig. 6. Thermal balance based on isolation concept

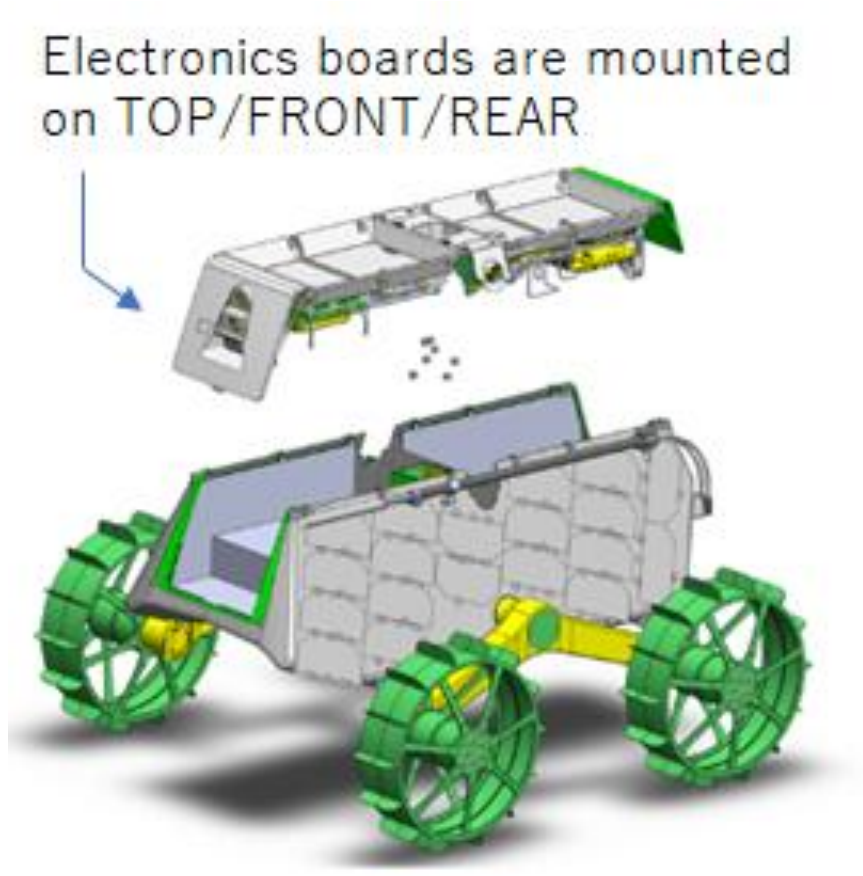

Fig. 7. Electronics layout concept

*2.3 Stepped Iterative Approach*

To ensure an agile and lean design cycle, we implemented an iterative design flow consisting of two phases: an analytic simulation phase for quick assessments and a numerical simulation phase for detailed analysis.

In the analytic simulation phase, we determined steady-state temperatures for each node by utilizing a spreadsheet-based parametric analytic model based on the equations discussed earlier in this paper (Eq 3-9). Using this information, we conducted a trade-off analysis involving the size of the top radiator, the angles of the side panels, and the angles of the rear and front surfaces to identify the optimal mechanical design for the rover.

In the subsequent numerical simulation phase, we performed finite element model simulations based on the rover concept developed during the analytic phase. This phase enabled us to calculate more precise heat exchanges that were intentionally not considered in the preliminary phase.

*2.4 Materials*

As we aimed to achieve a design with a small mass and high thermal performance, we explored the use of various industrial materials and conducted several evaluation tests on them, as listed in Table 3.

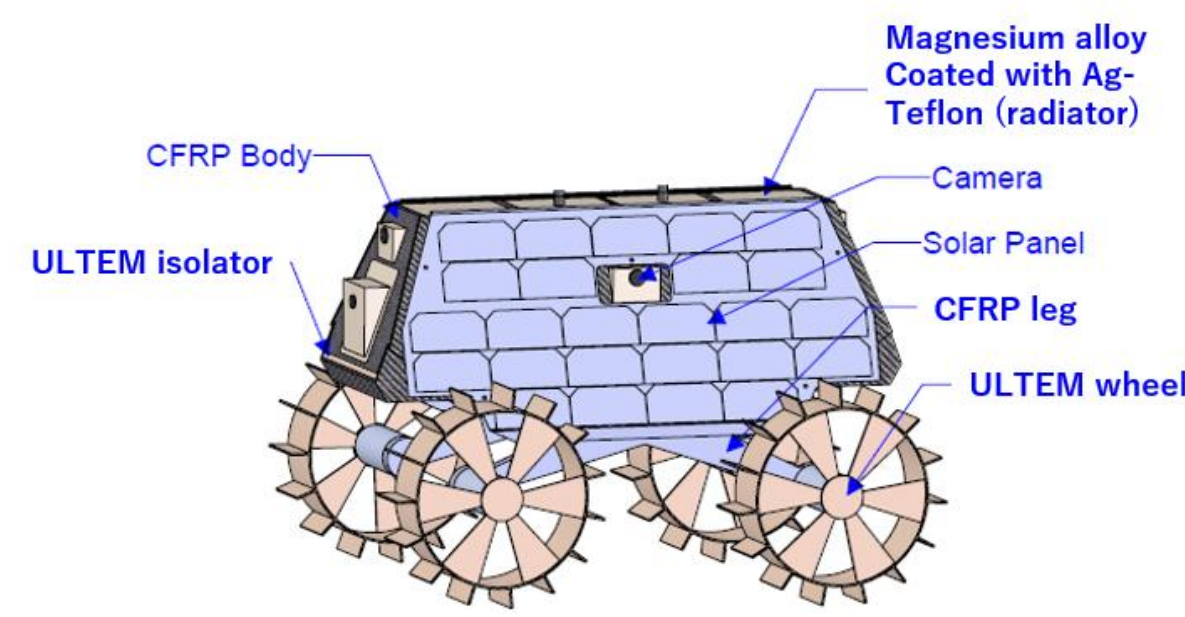

Fig. 8. Materials used in the HAKUTO rover

Table 3. Material selections

| Material | Part | Remarks |
|---|---|---|
| ULTEM | Isolator blocks in between panels, wheels | High thermal resistance, High thermal tolerance, high mechanical strength |
| Mg plate | Radiator | Light but good thermal conductivity |
| CFRP-Nomex Honeycomb | Body | Light mass, high mechanical strength, low thermal conductivity |
| Lamda Gel | Thermal connection between high power IC and Radiator | Flexible and high thermal conductance |
| Ag Teflon Tape | Radiator (Flat surface) | Low solar absorptivity and high emissivity |
| Ag Teflon Coating | Radiator (Curved surface) | Low solar absorptivity and high emissivity |

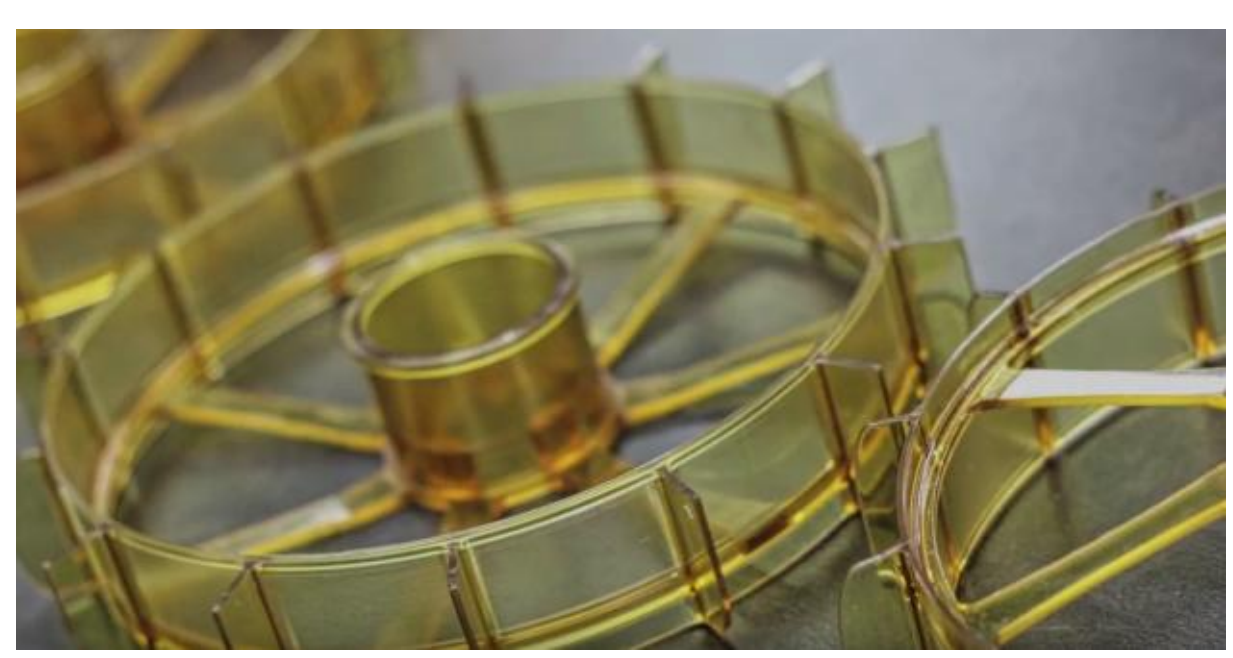
Fig. 9. ULTEM: industrial plastic that has high heat resistance, high heat tolerance, and high mechanical tolerance (used in wheels and thermal isolator blocks)

*2.5 Component Layout based on Classification*

In addition to the basic thermal design concepts discussed in previous sections of this paper, the requirement for a compact rover prompted us to further optimize the layout of the electronics board on the top radiator.

We divided the electronics components into two groups: thermal-sensitive components and thermally-tolerant components. The first group includes the battery and the radiation sensor, while the remaining electronics belong to the second group. These components were organized into temperature groupings known as the cold zone, where the temperature is controlled between -5°C and +40°C, and the hot zone, where the temperature ranges from -30°C to +65°C, as depicted in Fig. 10. Consequently, since we only needed to size the radiator to maintain the temperature within the specified limit for each zone, we were able to significantly reduce the size of the radiator. Figure 11 shows the actual electronics layout of the FM rover.

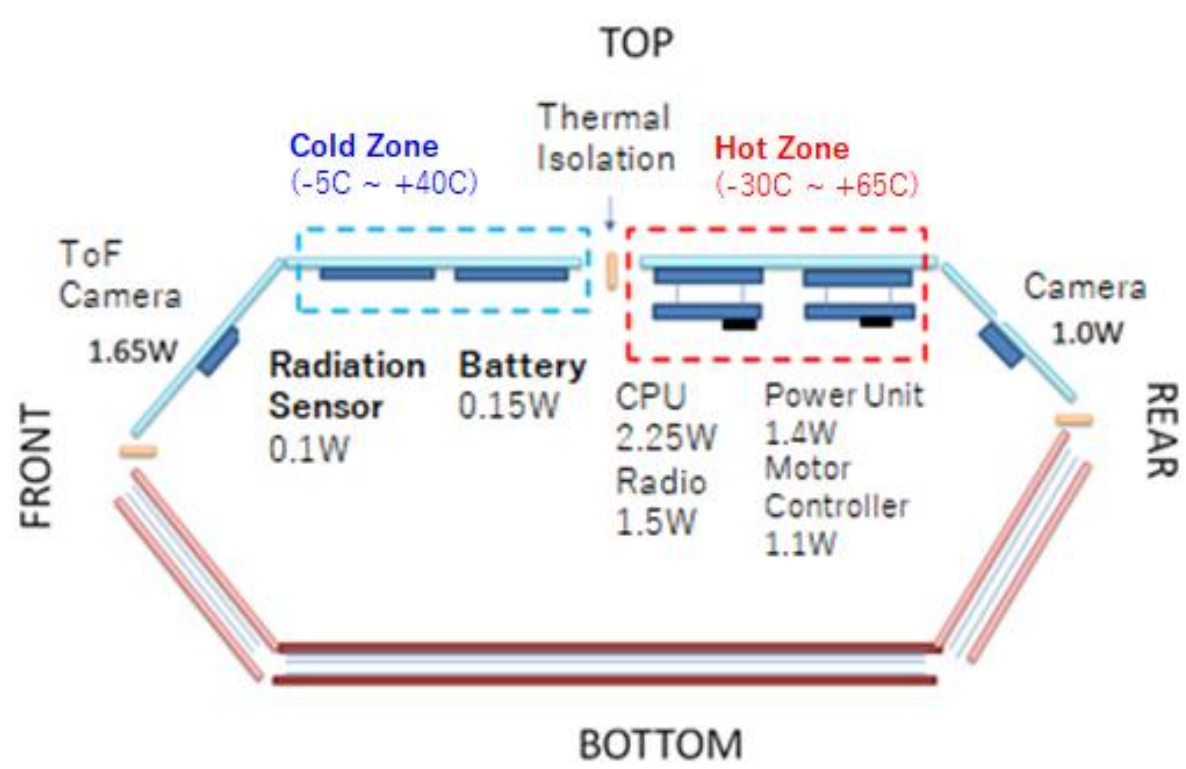

Fig. 10. Component layout based on hardware classification (cross-section from side view)

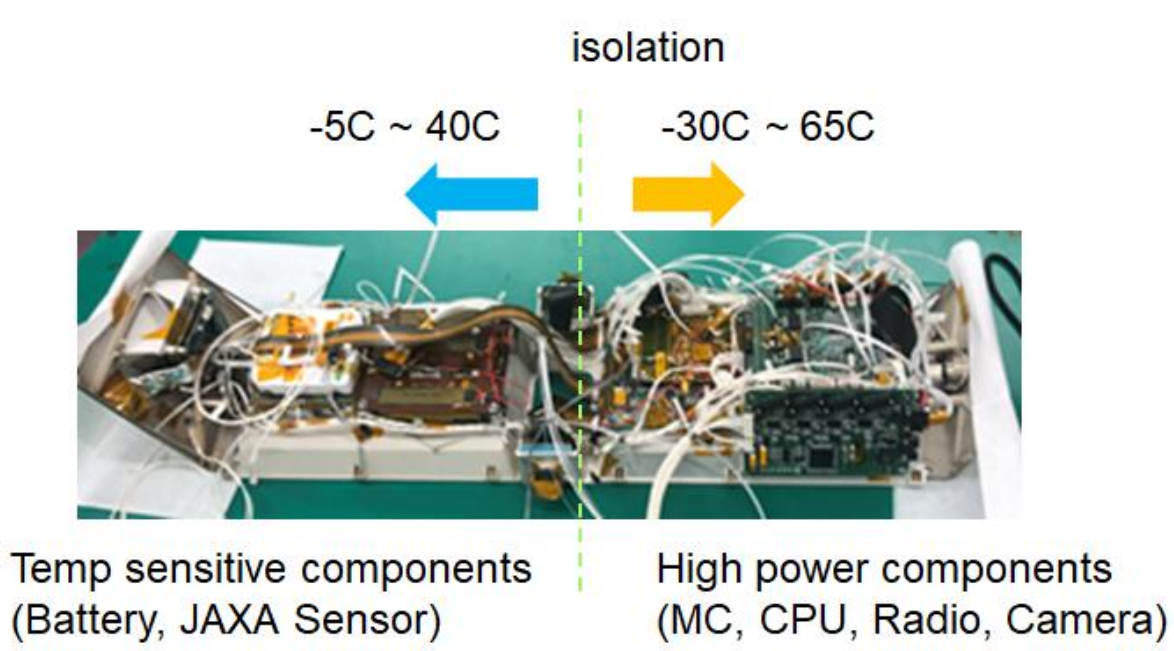

Fig. 11. A picture of electronics layout of the FM rover

*2.6 Indented Radiator*

In the HAKUTO mission, during the final design phase, the lunar latitude of the new landing site was changed to a lower value than initially planned, and the deployment timing of the new landing profile was advanced compared to the original plan set by the prime payload customer. As a result, the lunar surface temperature exhibited greater variations according to the new mission profile, with colder temperatures in the lunar morning and hotter temperatures at lunar noon. This necessitated an extension of the thermal design to accommodate a wider temperature range, while still maintaining the integrity of the main structure design to minimize impacts on the mechanical design and manufacturing process.

To address this challenge, we introduced a unique design approach known as the indented radiator. The indented radiator consists of two components: an angled surface painted black to enhance heat absorption and a flat surface covered with Ag-Teflon, as depicted in Fig.12. When the sun's inclination is low and the lunar surface temperature is also low, the system effectively absorbs heat through the angled surface. Conversely, when the sun's inclination angle is high and the lunar surface temperature is high, the system receives less sunlight. This design mitigates temperature variations experienced by the rover throughout the mission duration. However, it is important to note that this design will increase the maximum temperature experienced by the rover due to the black coating applied to the angled part, in contrast to its original state of being covered with Ag-Teflon and being flat (i.e., this design will reduce the temperature variations but increases the hottest temperature experienced by the rover).

Finally, we carefully designed the depth and angle of the indented structure to minimize temperature variations over time and ensure that the maximum temperature remains within the nominal operating range of the rover.

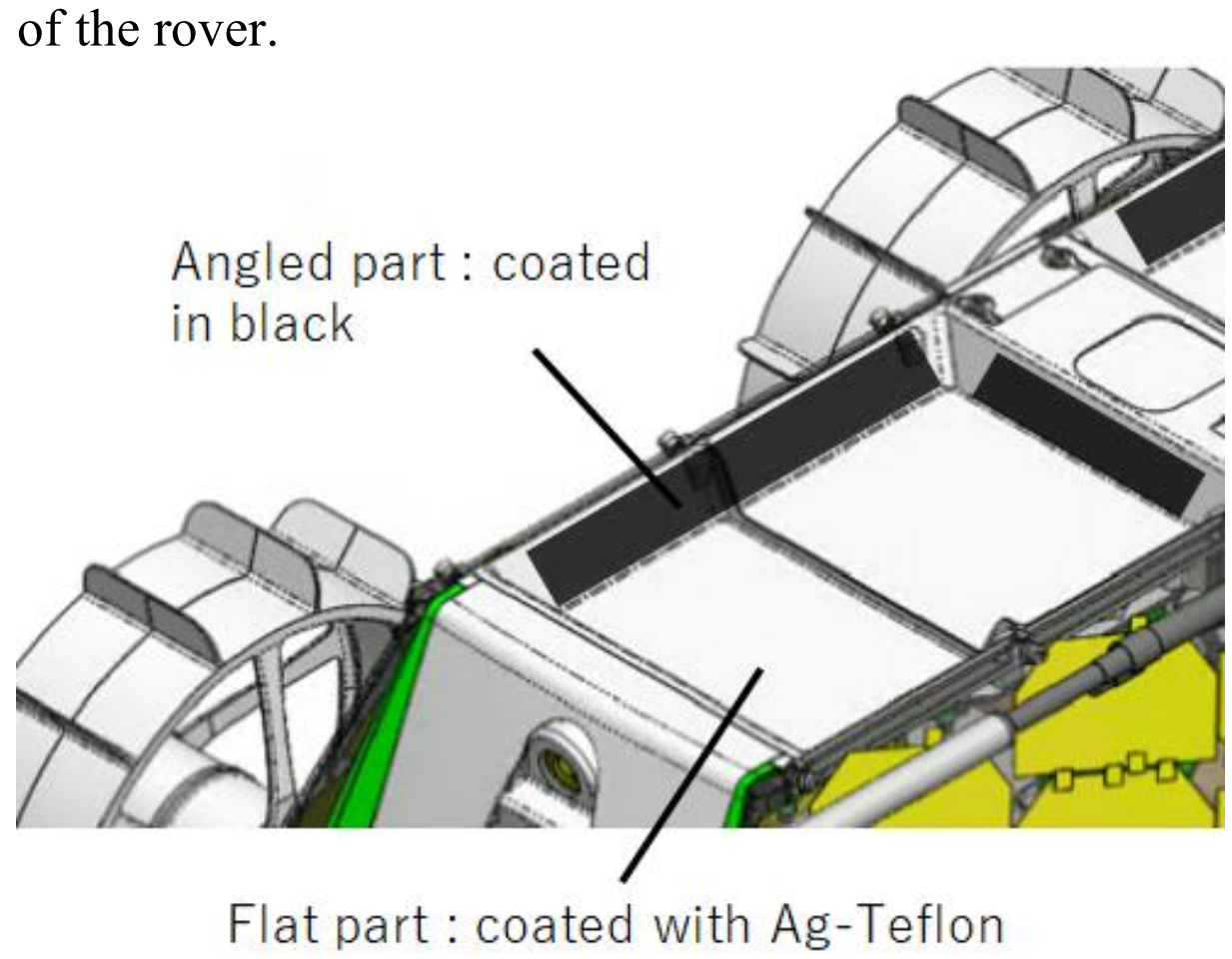

Fig. 12. Indented TOP Concept

*2.7 Safety Margin and Contamination Factor*

During the design phase, it is crucial to ensure that every component maintains a temperature margin of at least 10°C. Any deviation from this requirement should be identified and meticulously controlled. We conducted thermal simulations for each phase and verified that the simulation results satisfied the storage and operational temperature requirements for the components, with a margin of more than 10°C (except for the battery, which had a margin of 5°C).

During the analysis, we accounted for the potential impact of regolith contamination, which could cover up to 20% of the rover's radiator surface and solar cells. This contamination has the potential to degrade the heat dissipation performance.

## 3. Key Simulation Results

### *3.1 Cruise Phase*

We developed a simulation model of the cruise phase and identified the minimal heater power and margin necessary for the rover to survive in this phase.

### *3.2 Thrusting Phase*

We conducted transient analysis of the rover's temperature during thrust operations and confirmed that none of the components would exceed their storage temperature limits.

### *3.3 Deployment Phase*

After conducting a transient analysis of the rover's temperature during the deployment phase we found that the rover will have to move out of the shadow region **within an hour** of its deployment. This operational constraint has been accommodated in the operational plan.

### *3.4 Surface Operation Phase*

We performed simulation analysis for each condition outlined in Table 4. Through these simulations, we ensured that the rover remains within its operational temperature limits under all specified conditions. Additionally, we recognized the requirement for rover orientation control, in addition to heater control, during the lunar morning to warm up the rover. These specific details have been integrated into the operational and power budget designs to address this need.

Table 4. Surface Phase Simulation Conditions

| Case | Time in a day | Rover Orientation |
|---|---|---|
| 1 | 50hr after dawn | SIDE faces sun |
| 2 | 50hr after dawn | FRONT faces sun |
| 3 | 50hr after dawn | REAR faces sun |
| 4 | at lunar noon | SIDE faces sun |
| 5 | at lunar noon | FRONT faces sun |
| 6 | at lunar noon | REAR faces sun |

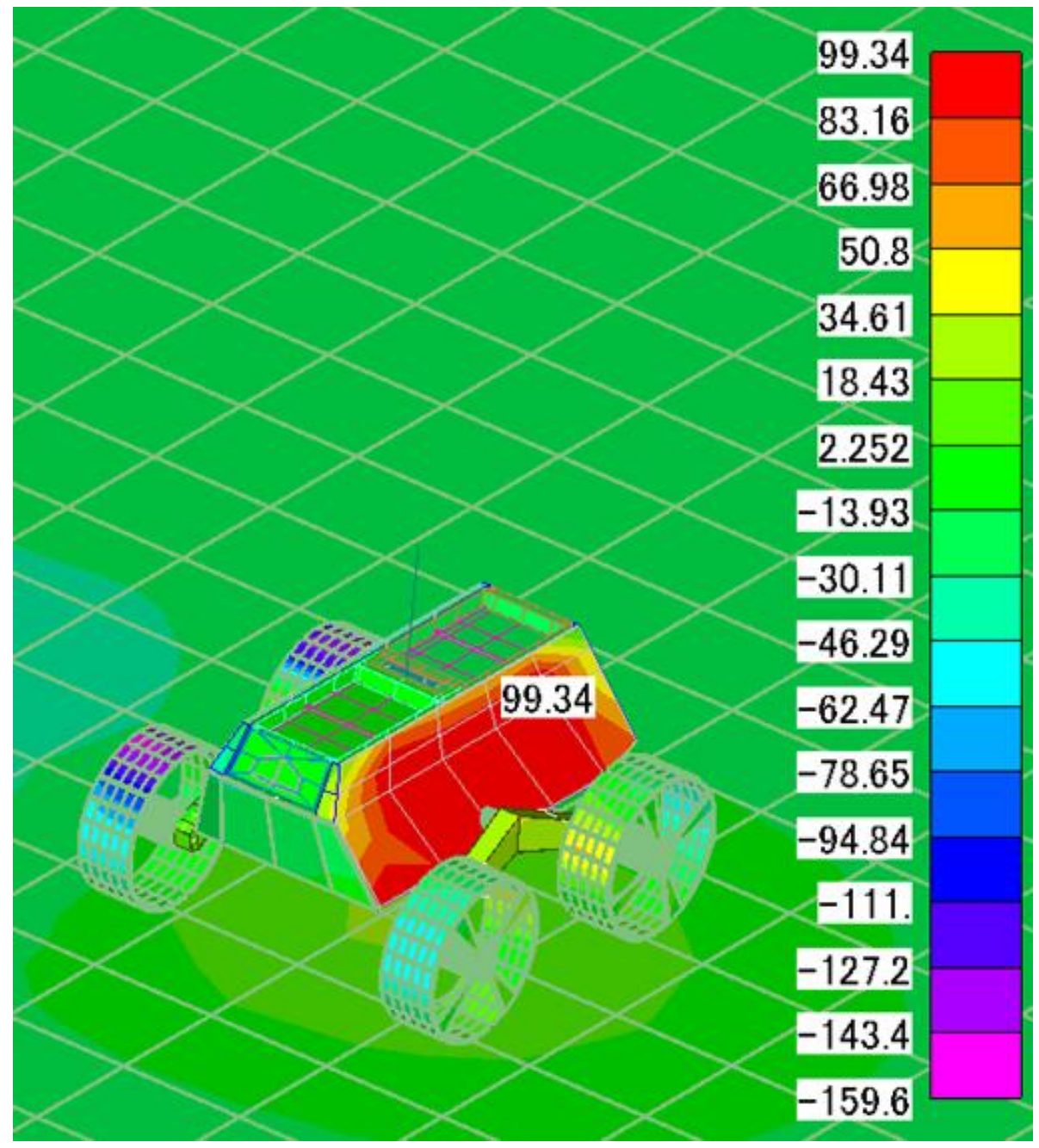

Fig. 13. Surface operation phase simulation

## 4. Validation Process

We included two levels of validation: unit-level material performance test and integrated thermal balance test.

### *4.1 Unit-Level Material Performance Tests*

We measured some thermal parameters at the unit level as presented in Table 5. The outcomes of these measurements were then integrated into the numerical simulation for further analysis and refinement.

Table 5. Parameters measured through Unit-Level Tests

| Parameter to measure | Methodology |
|---|---|
| ULTEM thermal contact conductance | Thermal conductimetry |
| Lamda Gel thermal contact conductance | Unit Thermal Vacuum |
| CFRP NOMEX Honeycomb material conductivity | Thermal conductimetry |
| Ag Teflon Coating absorptivity and emissivity | Optical measurement |
| ULTEM absorptivity and emissivity | Optical measurement |
| MLI effective emissivity | Unit Thermal Vacuum |

*4.2 Thermal Balance Test Simulating Cruise Phase*

To simulate the cruise thermal environment, we utilized a thermal vacuum chamber. The rover was securely mounted on a thermally controlled base plate, which was maintained at a temperature of -50°C to mimic the lowest temperature experienced on the lander deck during the cruise phase. Additionally, we exposed the rover to the cryogenic chamber wall, which was set to a temperature of -170°C, as depicted in Fig.14. More detailed thermal profile is presented in Fig. 15. Figure 16 shows the actual photography of the cruse environment test.

By comparing the measured temperatures with the simulation model, we verified that the temperature deviation of the mission-critical electronics components was within a margin of less than 5°C. This indicates that the thermal design successfully maintained the required temperature levels for these crucial components during the simulated cruise phase.

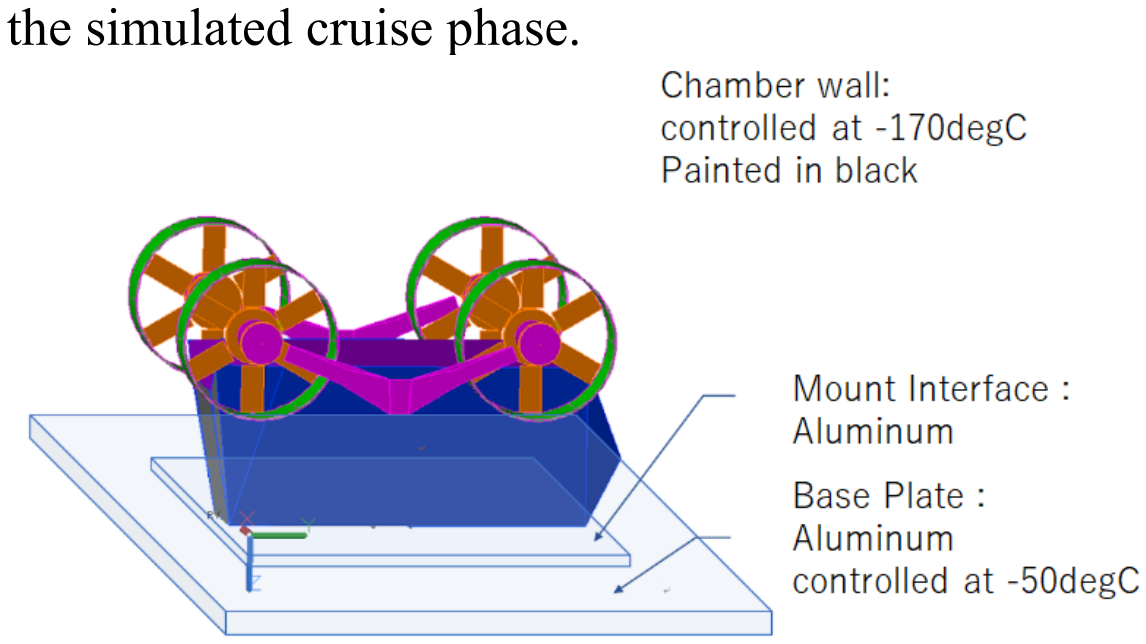

Fig. 14. Thermal balance test configuration emulating the cruise phase

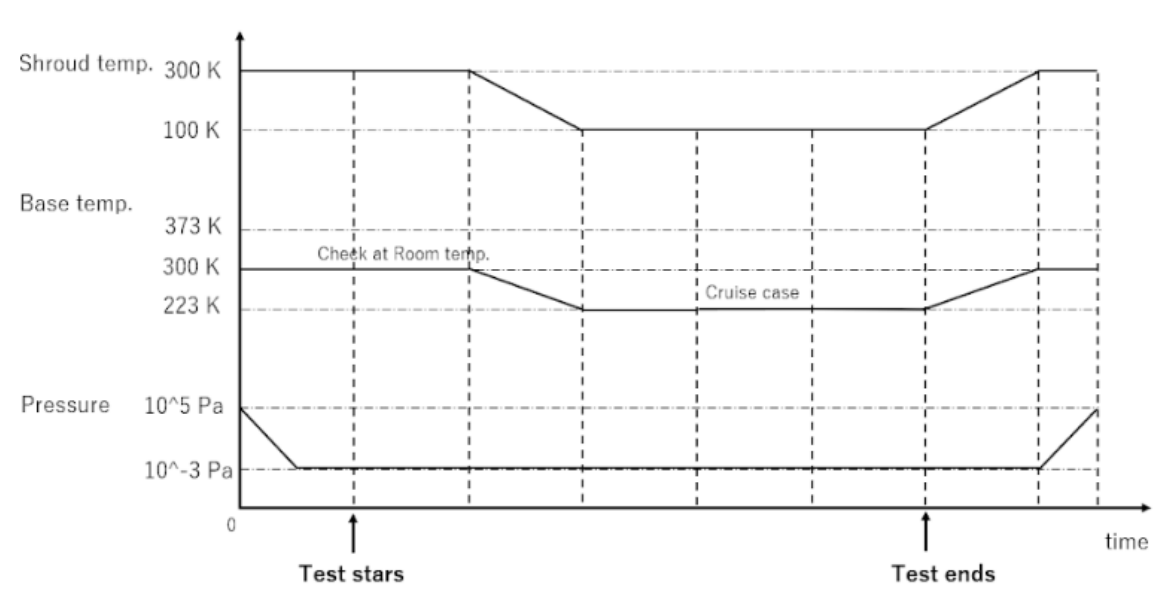

Fig. 15. Thermal profile used for the cruise phase test

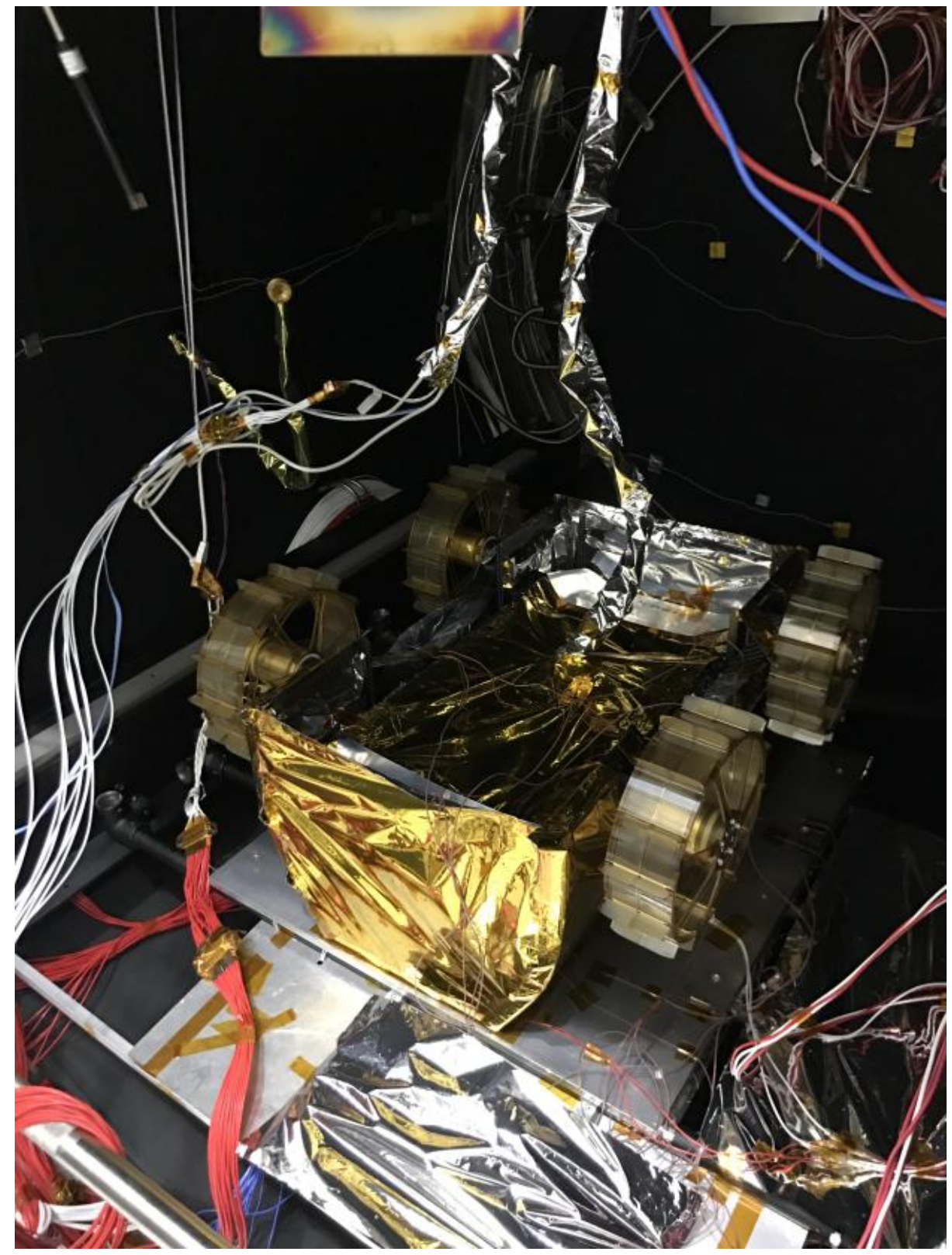

Fig. 16. A picture of the thermal balance test emulating the cruise phase. The rover was mounted onto thermally controlled plate simulating the lunar lander interface temperature during the cruise phase.

*4.3 Thermal Balance Test Simulating Surface Operation Phase*

To replicate the surface operating environment, we utilized a thermal vacuum chamber. The rover was positioned on a thermally controlled base plate, maintained at a temperature of 100°C, representing the highest temperature experienced on the lunar surface during the surface mission phase. Subsequently, the rover was heated using an IR light source, as illustrated in Fig. 17, with the IR light calibrated to raise the temperature of the rover's side to reach 100°C. This thermal condition was then replicated in the numerical simulation. More detailed thermal profile is presented in Fig. 18. Furthermore, Figure 19 displays an actual photograph of the cruise environment test.

By comparing the measured temperatures with the simulation model, we verified that the temperature deviation of the mission-critical electronics components was within a margin of less than 5°C. This demonstrates that the thermal design effectively maintained the required temperature levels for these essential components during the simulated surface mission phase.

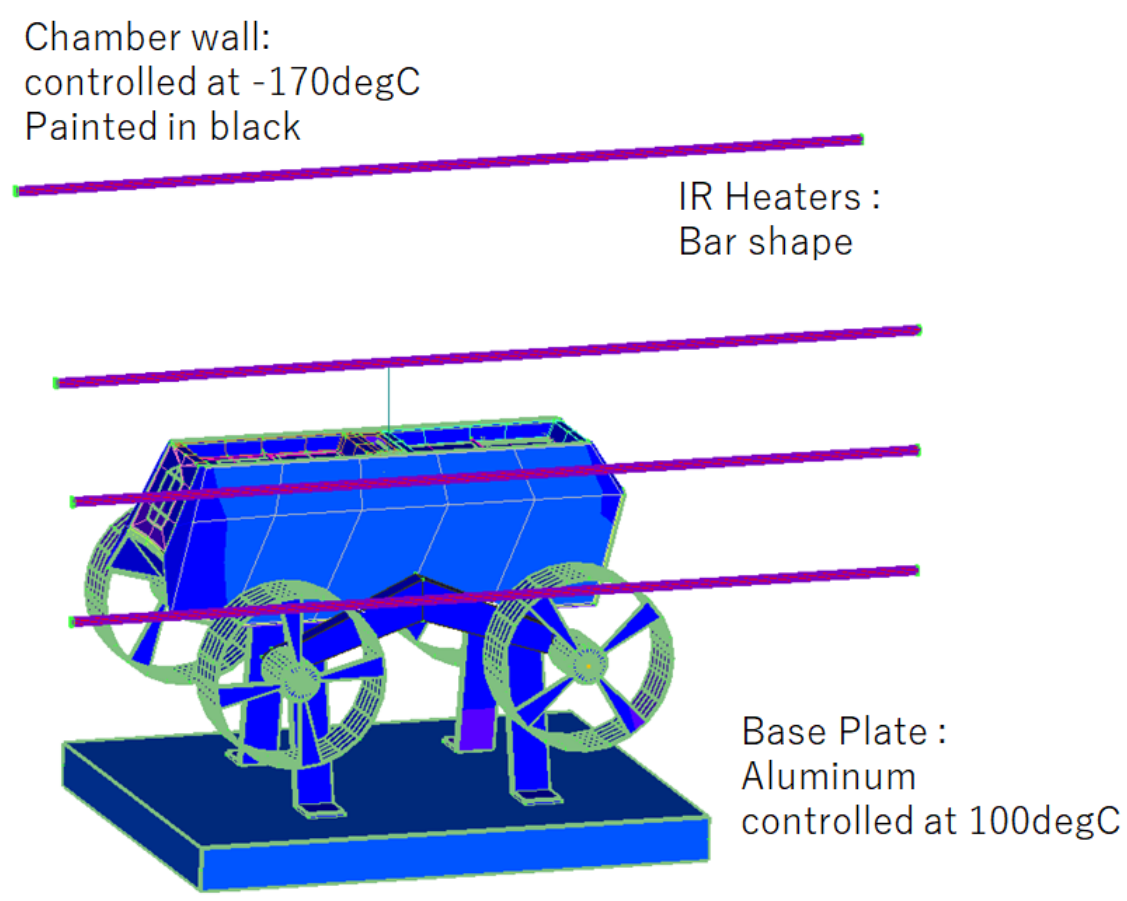

Fig. 17. Thermal balance test configuration emulating the surface mission phase

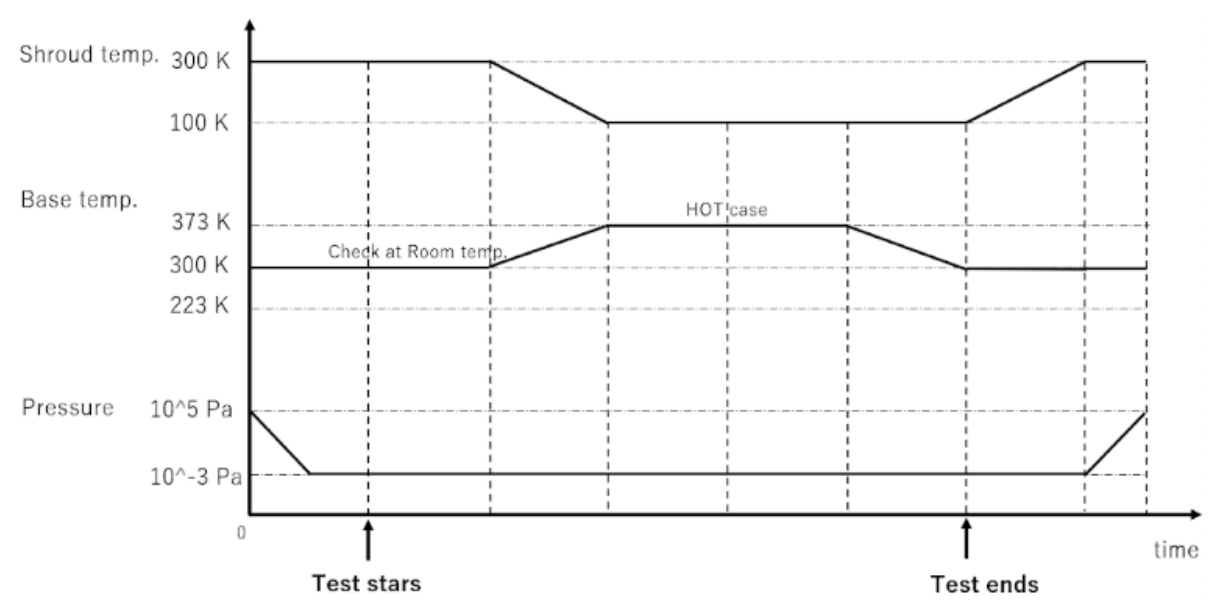

Fig. 18. Thermal profile used for the surface mission phase test

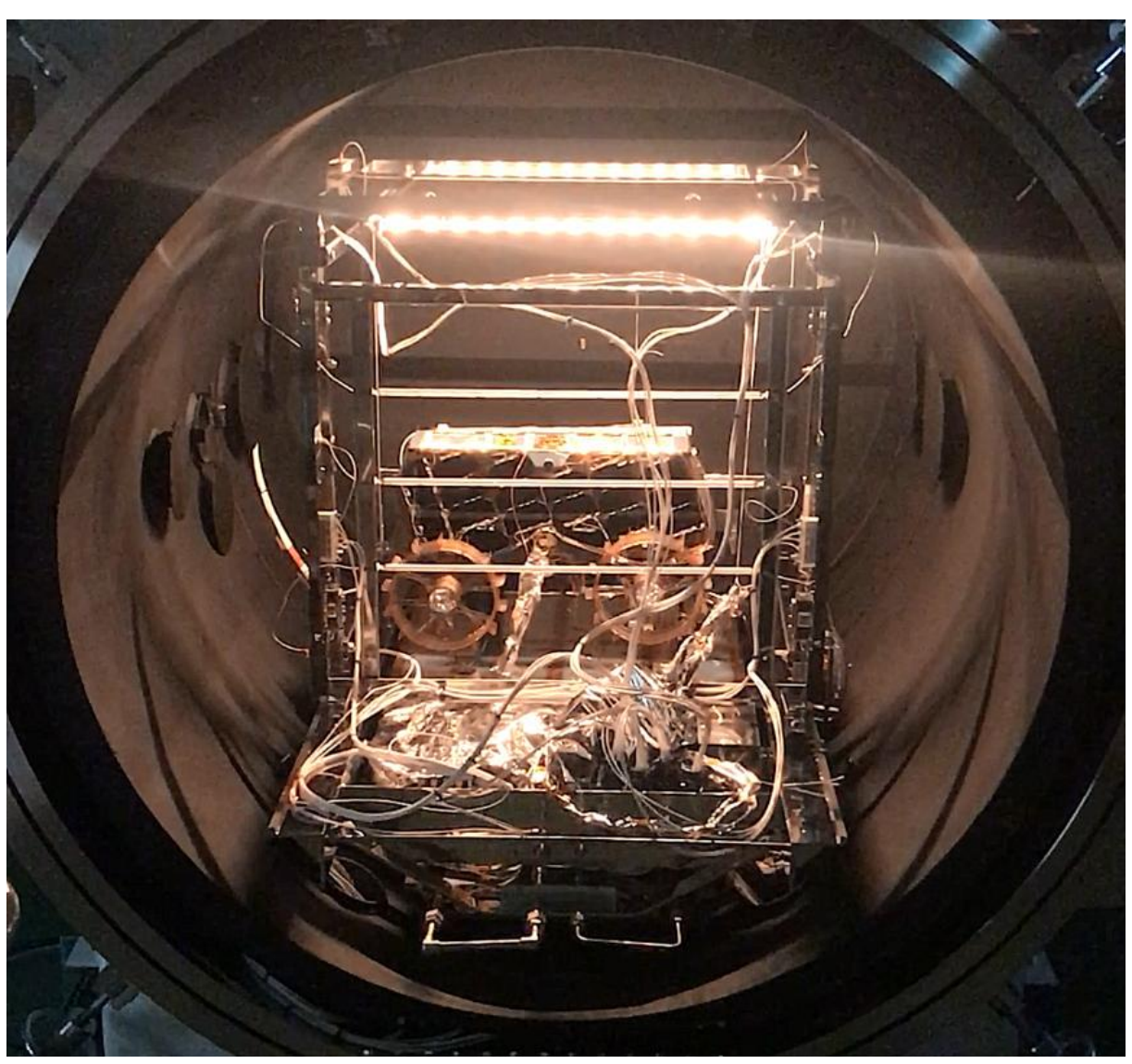
Fig. 19. A picture of the thermal balance test emulating the surface mission phase. The rover was mounted on the thermally controlled plate simulating the lunar surface temperature and lighted with IR light simulating the sunlight heat flux experience by the rover.

## 5. Discussion

Through simulation, we have confirmed that the design of the rover adheres to its specified operating and storage temperature ranges with a 10°C margin (except for the battery, which has a 5°C margin) throughout all mission phases. However, to meet the design requirements, the rover necessitates a heater and orientation control during the lunar morning.

Furthermore, a thermal balance test was conducted, which demonstrated that the actual thermal behavior of the rover aligns with the simulation model within a deviation of 5°C. This confirms the consistency between the simulated thermal performance and the actual performance of the manufactured rover.

Lastly, the final mass of the rover, including the interface to the lander deck, is less than 4 kg.

## 6. Conclusions

In this research, we have presented the thermal design process for HAKUTO, a micro-size lunar exploration rover. Achieving thermal performance while adhering to a 4 kg mass constraint posed a significant challenge. To address this, we implemented several unique design approaches, including the isolation concept, a component layout based on hardware classification, and the indented radiator. Additionally, we identified industrial materials that offer good thermal performance with minimal impact on mass.

To validate the integrity of our design, we implemented a strategic testing approach that encompassed a range of assessments. This included conducting unit-level material performance tests and two system-level thermal balance tests. By performing these tests, we were able to verify that the actual thermal behavior of the manufactured rover aligned closely with the simulated thermal performance. Specifically, we ensured that the temperature deviations for mission-critical electronics components remained within a 5°C margin.

By combining innovative design approaches, careful material selection, and comprehensive validation tests, we successfully achieved the thermal design objectives for the HAKUTO rover while maintaining its low mass.

## Acknowledgments

We would like to thank the all the members of HAKUTO Team for assisting the thermal development and all the corporate partners for providing technical advice as well as financial support.